\documentclass[sigconf]{acmart}
\def\BibTeX{{\rm B\kern-.05em{\sc i\kern-.025em b}\kern-.08emT\kern-.1667em\lower.7ex\hbox{E}\kern-.125emX}}
    
\copyrightyear{2019} 
\acmYear{2019} 
\setcopyright{acmcopyright}
\acmConference[DAC '19]{The 56th Annual Design Automation Conference 2019}{June 2--6, 2019}{Las Vegas, NV, USA\vspace{-2pt}}
\acmPrice{15.00\vspace{-2pt}}
\acmDOI{10.1145/3316781.3317829}
\acmISBN{978-1-4503-6725-7/19/06}

\usepackage[font=small]{caption}
\usepackage{subfigure}
\usepackage{verbatim}
\usepackage{amsmath}
\usepackage{amssymb}
\usepackage{booktabs} 
\usepackage{algorithm}
\usepackage{algcompatible}
\usepackage{algpseudocode}
\usepackage{multirow}
\usepackage{color}
\usepackage{threeparttable}
\usepackage{pifont}

\usepackage{tikz}
\usepackage{calc}

\begin{document}

%
\title{
\vspace{-0pt} FPGA/DNN Co-Design: An Efficient Design Methodology for IoT Intelligence on the Edge \vspace{-8pt}
}

\affiliation{\large 
Cong Hao$^{1*}$, Xiaofan Zhang$^{1*}$, Yuhong Li$^1$, Sitao Huang$^1$, Jinjun Xiong$^2$, Kyle Rupnow$^3$, Wen-mei Hwu$^1$, Deming Chen$^{1,3}$}

\affiliation{ \normalsize \vspace{-10pt}
	\institution{$^1$University of Illinois at Urbana-Champaign, $^2$IBM T. J. Watson Research Center, $^3$Inspirit IoT, Inc.}
}
\affiliation{ \normalsize
	\textit{\{congh, xiaofan3, leeyh, shuang91, w-hwu, dchen\}@illinois.edu, jinjun@us.ibm.com, kyle.rupnow@inspirit-iot.com}
}

\begin{abstract}
\vspace{-2pt}
While embedded FPGAs are attractive platforms for DNN acceleration on edge-devices due to their low latency and high energy efficiency,
the scarcity of resources of edge-scale FPGA devices also makes it challenging for DNN deployment.
In this paper, we propose a simultaneous FPGA/DNN co-design methodology with both bottom-up and top-down approaches:
a bottom-up hardware-oriented DNN model search for high accuracy, and a top-down FPGA accelerator design 
considering DNN-specific characteristics.
We also build an automatic co-design flow, including an \textit{Auto-DNN} engine to perform hardware-oriented DNN model search,
as well as an \textit{Auto-HLS} engine 
to generate synthesizable C code of the FPGA accelerator for explored DNNs.
We demonstrate our co-design approach on an object detection task using PYNQ-Z1 FPGA. Results show that our proposed DNN model and accelerator outperform the state-of-the-art FPGA designs in all aspects including Intersection-over-Union (IoU) (6.2\% higher), frames per second (FPS) (2.48$\times$ higher), power consumption (40\% lower), and energy efficiency (2.5$\times$ higher). Compared to GPU-based solutions, our designs deliver similar accuracy but consume far less energy.
\end{abstract}

%
\maketitle

\vspace{-2pt}

\section{Introduction}
\vspace{-2pt}
The world has seen rapid adoption of FPGAs for DNN acceleration  \cite{zhang2015optimizing,qiu2016going,Xiaofan2017High,junsong2018elb,zhang2018dnnbuilder,li2019implementing}.
Internet of Things (IoT) applications in domains such as self-driving, security and surveillance face particular challenges as they require both
sophisticated DNN models for Quality of Results (QoR) and strict latency, power, and resource constraints. 
Embedded FPGAs are one of the most attractive candidates to enable machine learning capability for IoT applications~\cite{zhang2017machine} because of their high energy efficiency and low cost,
but the scarcity of resources also makes DNN accelerator design and deployment on FPGA more challenging.
In a typical top-down design flow, DNN models are first designed concentrating more on the QoR,
expecting the accelerator can meet performance constraints through later optimization.
This approach has been largely successful,
but ignores the impact that deployment architecture should have on the DNN design.
Instead, DNNs should be built bottom-up with adequate understanding of the hardware constraints before expanding network size to reach the targeted QoR. 
Most importantly, DNNs and the corresponding FPGA accelerators need to be developed simultaneously, and we believe in FPGA/DNN co-design as a promising solution with immense optimization opportunity: 
DNN designs should be FPGA-architecture driven, and FPGA accelerators should be DNN-aware. 

Despite the opportunities, a good co-design approach requires the exploration of an extremely large number of variables in the combined DNN and FPGA accelerator co-design space, and constrains the solutions to have both high QoR and efficient FPGA implementations.
Consequently, the co-design task will be extremely time-consuming, as we must perform training of each candidate DNN to evaluate its quality.
Even using Neural Architecture Search (NAS)~\cite{zoph2017learning, zoph2016neural} for DNN development and the High Level Synthesis (HLS) for fast FPGA development~\cite{chen2010lopass, rupnow2011high},
both tasks still need a large amount of engineering hours.

Facing the opportunities and challenges, in this work, we propose a simultaneous FPGA/DNN co-design approach,
which effectively searches the design space to both generate high quality DNNs suitable for FPGA deployment, and highly optimized FPGA accelerators.
The contributions are summarized as follows:
\leftmargini=4mm
\begin{itemize}
\item {
\vspace{-4pt}
We propose the first simultaneous FPGA/DNN co-design methodology with (1) hardware-oriented DNN model design following bottom-up approach, and (2) DNN-driven FPGA accelerator design following top-down approach. A fully automatic co-design flow is developed accordingly for simultaneous DNN search and FPGA accelerator generation.
}
\item{
For DNN model design, we introduce a DNN template to guide the DNN generation with predictable performance and resource utilization, which greatly reduces the co-design search space.
Based on such template, an automatic DNN model search engine, \textit{Auto-DNN},
is proposed to effectively explore the design space and generate DNN models for desired QoR.
}

\item{
For FPGA accelerator design, we introduce a fine-grained tile-based pipeline architecture, which supports arbitrary DNNs generated by \textit{Auto-DNN} using a library of highly optimized HLS IPs.
Based on such architecture, an automatic HLS generator, \textit{Auto-HLS}, is proposed to directly generate synthesizable C code of the DNN models,
to conduct latency/resource estimation and FPGA accelerator generation.
}

\item{
We demonstrate our co-design approach on an object detection task targeting a PYNQ-Z1 embedded FPGA.
DNN models are searched and mapped to the board 
with the state-of-the-art performance regarding accuracy, speed, and power efficiency.
}
\end{itemize}
\vspace{-8pt}

\vspace{-2pt}
\section{Related Work}
\vspace{-2pt}

DNN model design and FPGA accelerator design are each under intense study, but
these activities are often conducted independently. DNN design is conducted
either manually by machine learning experts or automatically by Neural
Architecture Search (NAS) such as recursive neural networks (RNN)~\cite{zoph2017learning} and reinforcement learning~\cite{real2018regularized}.
Although high QoR can be obtained, the DNNs may have complex structures that are unsuitable for FPGA deployment.
A few platform-aware DNN search methods are proposed, such as~\cite{tan2018mnasnet, cai2018proxylessnas},
but they only consider the DNN inference latency on CPUs and GPUs, not on FPGAs.
On the other hand, for FPGA-based DNN accelerator, recent technologies such as quantization~\cite{qiu2016going,junsong2018elb} and model compression~\cite{han2017ese} are used to reduce DNN model size, 
and latency-directed resource allocation~\cite{Xiaofan2017High} and fine-grained pipeline architecture~\cite{zhang2018dnnbuilder} are proposed to deliver low latency during DNN inference.
However, these approaches may be limited by the DNN models,
and may not have sufficient optimization opportunities to meet performance constraints on target IoT platforms.
Other works specifically conduct design space exploration to select hardware configuration parameters~\cite{motamedi2016design,zhong2017design} together with optimizations including loop unrolling and pipelining, but they do not explore configurations on the DNN side, which could make hardware implementations more effective.



\begin{figure*}[t]
\centering
\includegraphics[width=0.98\textwidth]{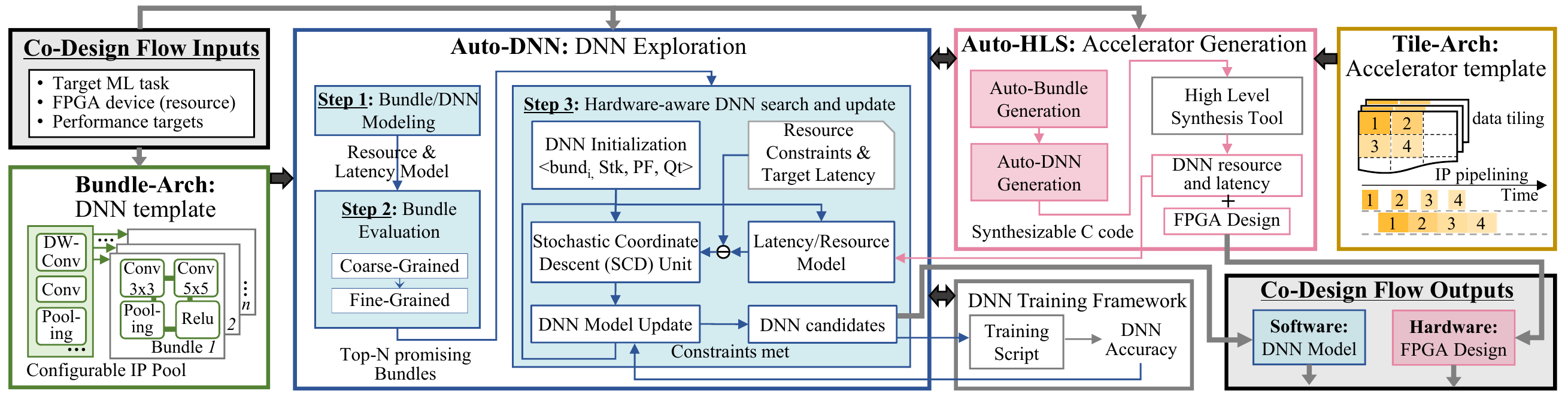}
\vspace{-10pt}
\caption{The overall FPGA/DNN co-design flow is composed of four key components: \textit{Bundle-Arch} as a hardware-aware DNN template (green); \textit{Auto-DNN} for DNN exploration (blue); \textit{Auto-HLS} for FPGA accelerator synthesizable C code generation (pink); \textit{Tile-Arch} as a low-latency accelerator template (yellow). Auto-DNN works as the primary component and outputs DNN models, while Auto-HLS outputs the corresponding FPGA implementations of the DNN models.}
\label{fig:overall_flow}
\end{figure*}

\vspace{-4pt}
\section{FPGA/DNN Co-design}
\vspace{-2pt}


\subsection{Co-Design Space}
There is a large design space for DNN design, such as the number and types of layers, the number of input/output channels, residual connections, concatenations, etc.
Similarly, the design space for FPGA accelerator is also enormous, such as IP instance categories, IP reuse strategies, quantization schemes, parallel factors, data transfer behaviors, and buffer sizes, etc.
Thus, to cover both DNN model and accelerator design, the co-design space is exponentially greater than any of the above, which requires effective techniques to find high quality solutions.
In this work, we conduct FPGA/DNN exploration by proposing a \textit{co-design space} to efficiently narrow down the effort for space searching.

The variables in the proposed co-design space are summarized in Table~\ref{tab:co-design-var}.
For FPGA accelerator, we use IP-based design strategy as in~\cite{Xiaofan2017High,zhang2018dnnbuilder}.
Each IP supports a basic DNN layer type (e.g. Conv, Pooling), which must be instantiated and configured if the DNN model contains such type of layer.
$L$ is the total number of DNN layers.
$IP_1$ to $IP_m$ represent the available configurable IP templates.
$p_j(1 \leq j \leq n)$ represent the configured IP instances,
where the configurable parameters include
parallelism factor $PF_j$ and quantization scheme $Q_j$.
<$l_j^1, \cdots l_j^z$> represents the layers for which an IP instance $p_j$ is used in FPGA to conduct the computation.
Vector <$f_{ch_1}, f_{ch_2}, \cdots, f_{ch_{L}}$> represents the expansions of channel depth through the entire DNN.
In addition, $ds_1$ to $ds_k$ represent down sampling layers with a down sampling factor $f_{ds_i}$.
The combination of these parameters can specify both the DNN model and the accelerator design.

\begin{table}[t]
\footnotesize
\centering
\caption{Key Variables for FPGA/DNN Co-Design\vspace{-8pt}} \label{tab:co-design-var}
\renewcommand{\arraystretch}{0.9}
\setlength{\tabcolsep}{1.8pt}
\begin{tabular}{|c|c| c |}
\hline
Variables &  Explanation & Effect\\ \hline
$L$ & Total number of layers & A, P, R\\
$IP_1, IP_2, \cdots, IP_m$ & IP templates for DNN building & A, P, R\\ 
$p_1, p_2, \cdots, p_n$ & Labels for IP instances & P, R\\ 
$\langle PF_j, Q_j\rangle$ & Configuration for $p_j (1 \leq j \leq n)$ & A, P, R\\
$\langle l_j^1, \cdots, l_j^z\rangle$ & The layers where $p_j$ is used & A, P\\
$<f_{ch_1}, f_{ch_2}, \cdots, f_{ch_{L}}>$ & Channel expansion factors & A, P, R \\
$ds_1, ds_2, \cdots, ds_k$ & Down-sampling layers & A, P, R\\
$f_{ds_i}$ & Down-sampling factor & A, P, R\\
\hline 
\multicolumn{3}{|l|}{A: Accuracy, P: Performance, R: Resource} \\
\hline
\end{tabular}
\vspace{-16pt}
\end{table}

\vspace{-6pt}
\subsection{Overall Co-Design Flow}
\vspace{-2pt}

\begin{figure}[t]
\centering
\includegraphics[width=0.44\textwidth]{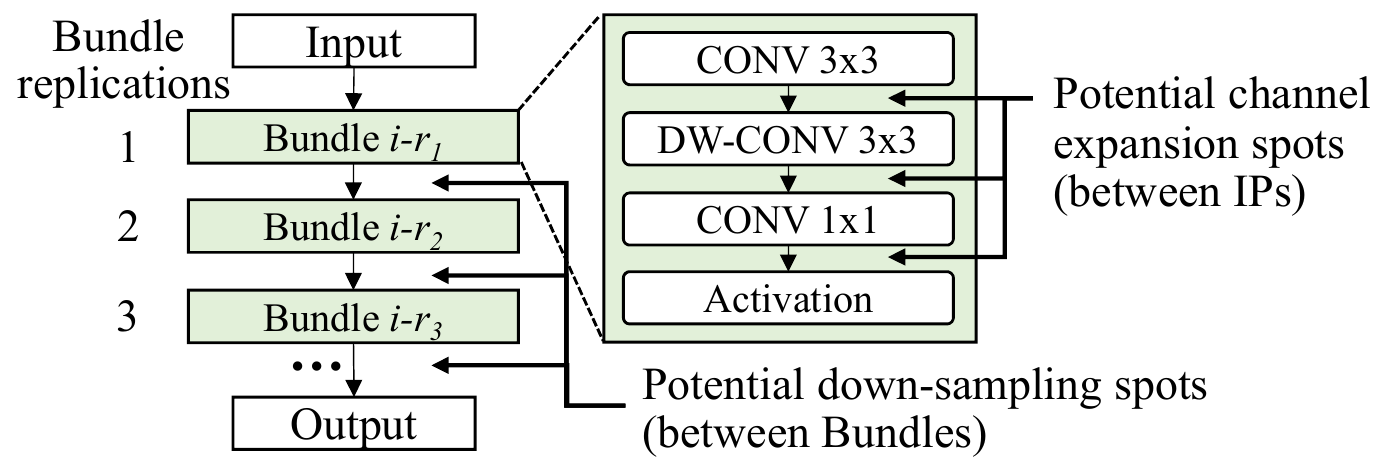}
\vspace{-8pt}
\caption{Bundle-Arch: A DNN template using pre-designed hardware-aware Bundles.}
\vspace{-12pt}
\label{fig:Bundle}
\end{figure}

Our co-design flow solves two design problems simultaneously:
the bottom-up DNN model exploration, and the top-down FPGA accelerator generation.
For DNN models, we start from basic hardware-aware building blocks, and gradually construct DNNs to reach desired QoR;
for FPGA accelerators, we follow a fixed architecture, and optimize configurable parameters to pursue most efficient DNN implementations.
%
Regarding the two tasks, we propose the following four key components:
\begin{itemize}
    \item {
For DNN: (1) \textbf{Bundle-Arch}: a hardware-aware DNN building block template to build up DNN models;
(2) \textbf{Auto-DNN}: an efficient search engine to explore DNN candidates under hardware resource and performance constraints;
    }
    \item {
For FPGA: (3) \textbf{Tile-Arch}: a low-latency FPGA accelerator template for DNN implementation;
(4) \textbf{Auto-HLS}: a fast board-level design generator to automatically map DNNs onto FPGAs.
    }
\end{itemize}

These four components work seamlessly as:
\textit{Auto-DNN} generates DNNs using \textit{Bundle-Arch} DNN templates,
while \textit{Auto-HLS} builds accelerators following the \textit{Tile-Arch} FPGA template. Meanwhile, \textit{Auto-DNN} and \textit{Auto-HLS} execute iteratively for DNN model search and FPGA accelerator generation.

%
Fig.~\ref{fig:overall_flow} shows the overall flow of our proposed co-design methodology, composed of the four key components.
The inputs include: targeted machine learning task (e.g., classification, detection), target FPGA device with resource constraints (e.g., DSP, LUTs, memory), and the performance targets of the accelerator (e.g., latency).
We also have configurable IP templates as inputs.
The outputs include hardware-oriented DNN models and their FPGA accelerators.
There are three major steps in our co-design flow:
\begin{enumerate}
    \item {\textbf{Co-Design Step 1: Building block and DNN modeling}.
    Given DNN building blocks and hardware IP pool, we first construct analytical models to capture the hardware latency and resource utilization of the building blocks and the DNNs built from the blocks. This is to provide performance estimation in the early stage of DNN exploration.
    }
    \item {
    \textbf{Co-Design Step 2: Building block selection}.
   To select the most promising DNN building blocks for the specific machine learning task and target FPGA, \textit{Auto-DNN} performs both coarse- and fine-grained evaluations of the building blocks regarding three most important features: latency, resource utilization and accuracy.
   Based on the evaluation, building blocks on the Pareto curve will be selected for further DNN exploration.
    }
    \item{
    \textbf{Co-Design Step 3: Hardware-aware DNN search and update}. 
    Given selected building blocks, \textit{Auto-DNN} explores the DNNs under given resource and latency constraints by using stochastic coordinate descent (SCD).
    DNNs output from SCD are passed to \textit{Auto-HLS} to get more precise performance and resource results, and are fed back to SCD for update.
    The generated DNNs that meet performance and resource requirements are output for training and fine-tuning.
    }
\end{enumerate}

 
In the following, the \textit{Bundle-Arch} and \textit{Tile-Arch} templates are introduced in Sec.~\ref{sec:templates}. Building block evaluation and DNN search using \textit{Auto-DNN} and \textit{Auto-HLS} are introduced in Sec.~\ref{sec:auto-dnn}.

\vspace{-4pt}
\section{DNN and Accelerator Template} \label{sec:templates}

\subsection{\large Bundle-Arch: Hardware-Aware DNN Template}
\label{sec:bundle-template}
We use DNN templates for model exploration because:
(1) they help narrow down the DNN design space and speedup the search process, and
(2) they can integrate hardware knowledge and guide DNN design towards hardware-oriented directions. 

We propose a hardware-aware \textit{Bundle} based DNN template, \textit{Bundle-Arch}.
A Bundle is a set of sequential DNN layers as a basic DNN building block. 
For example, a Bundle \textit{i-$r_1$} in Fig. \ref{fig:Bundle} contains four DNN layers cascaded from top to bottom.
DNN models are built by replicating, shaping and configuring a Bundle in a bottom-up manner.
In Fig.~\ref{fig:Bundle}, three replications of the same Bundle are shown,
where each replication may vary in input/output data dimensions.
Between Bundles, we reserve down-sampling spots for feature map size compression.
When implemented on FPGA, a hardware Bundle also represents a combination of the IP instances used for DNN layer computation.
The IPs within one Bundle are organized based on our proposed \textit{Tile-Arch} (in Sec.\ref{sec:pipeline-arch}), which delivers optimized low-latency designs.

We adopt the Bundle based strategy for building hardware-oriented DNNs following the same trend of modern popular DNNs,
such as the residual block in ResNet \cite{he2016deep}
and depth-wise blocks in Mobilenet \cite{sandler2018mobilenetv2}.
Moreover, FPGA accelerators can also benefit from pre-designed and optimized hardware Bundles, which provide more predictable patterns on computation and memory access behaviors.


\vspace{-8pt}
\subsection{\large Bundle Generation}
\vspace{-2pt}
To generate DNN Bundles,
we select the following IPs (DNN layers) similar to previous NAS works as:
convolution (conv) $1\times1$, $3\times3$, $5\times5$; depth-wise conv $3\times3$, $5\times5$, $7\times7$; max/avg pooling; 
normalization; activation.
In FPGA implementation, each IP requires at least one instance, and more IPs mean more resource overhead. 
In this work, we limit up to two computational IPs in each Bundle since we are targeting IoT devices with scarce resources. It can be easily extended to support more IPs for devices with more resources.

In our experiments, 18 Bundle candidates are generated offline and used for DNN exploration. However, as we have more IPs, the number of Bundles may grow significantly.
For scalability, the Bundles will be evaluated first (in Sec.~\ref{sec:bund-eval}), and the most promising ones will be selected for further DNN exploration based on their potential accuracy contributions and hardware characteristics.

\vspace{-8pt}
\subsection{\large Tile-Arch: Low Latency Accelerator Template \label{sec:pipeline-arch}}
\vspace{-2pt}

We propose a fine-grained tile-based pipeline accelerator architecture template, \textit{Tile-Arch},
for mapping DNNs onto embedded FPGAs, which can deliver low latency designs and exploit maximum resource saving.
This template has the following features:
\begin{itemize}
\item {
\vspace{-4pt}
Layer-level IP reuse: we adopt a folded overall structure, where the DNN layers are computed sequentially on FPGA by reusing IP instances across layers.
It can maximally exploit resource reuse, which is especially crucial for embedded FPGAs.
}
\item {
Tile-level IP reuse: resulting from layer-level IP reuse, the intermediate data between layers are partitioned into tiles of common size across all layers, and an IP instance is reused for multiple tiles. It allows direct data transfer between IP instances of subsequent layers without on-/off-chip memory access.
}
\item {
Tile-level pipelining: since data tiles within a layer do not have data dependencies, we can leverage tile-level IP pipelining both within a layer and across consecutive layers.
}
\vspace{-2pt}
\end{itemize}

\begin{figure}[t]
\centering
\vspace{-12pt}
\includegraphics[width=0.4\textwidth]{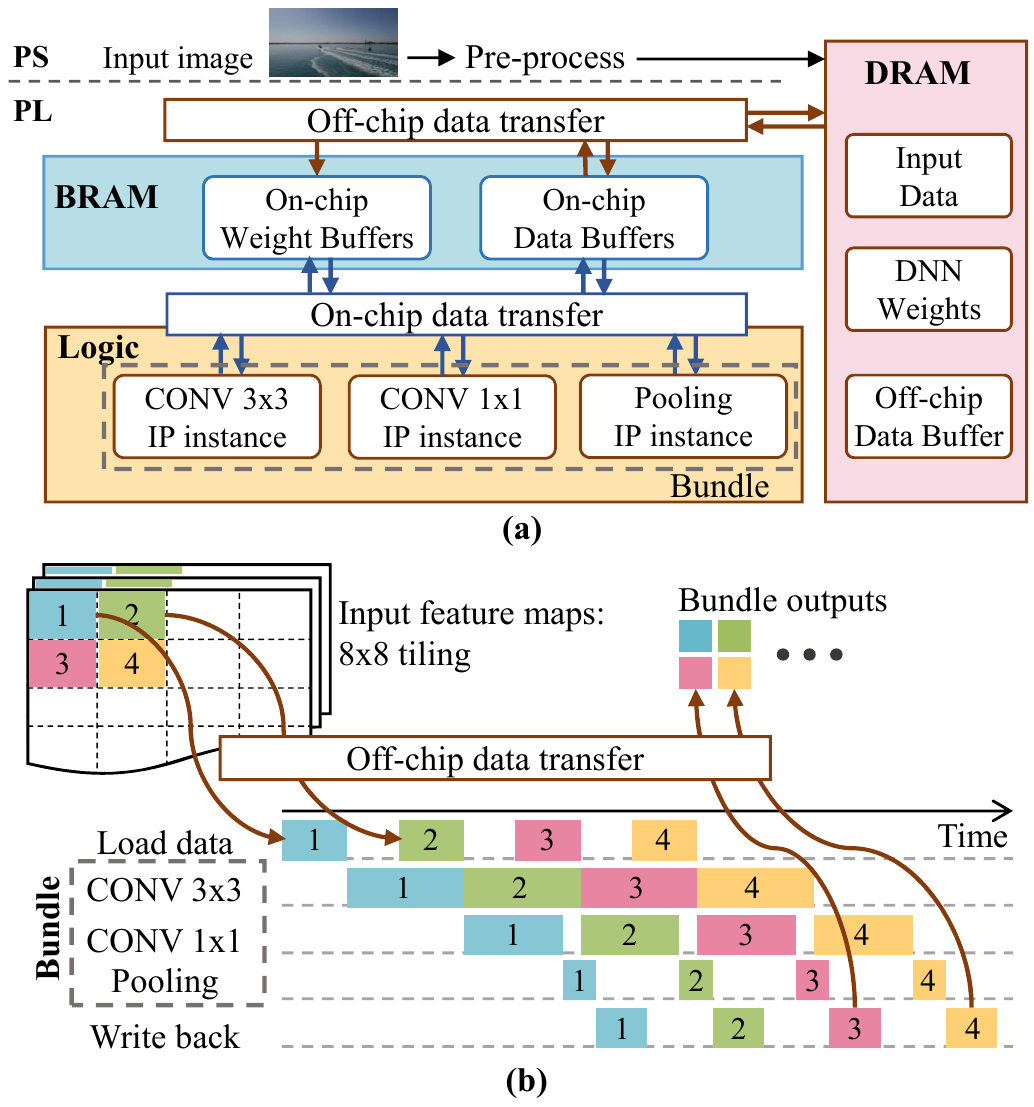}
\vspace{-12pt}
\caption{Tile-Arc: a low latency FPGA accelerator template with (a) a top-level diagram of the proposed architecture and (b) an example of tile-based pipeline structure.}
\vspace{-12pt}
\label{fig:overall_arch}
\end{figure}

Fig.~\ref{fig:overall_arch} (a) shows an example of the top-level diagram of the proposed template architecture.
In this example, the Bundle contains IP instances including conv $3\times3 $, $1\times1$ and pooling.
On-chip data buffers are allocated in BRAM for intra-Bundle communication,
while off-chip data buffers are allocated in DRAM for inter-Bundle communication.
Fig.~\ref{fig:overall_arch} (b) illustrates the tile-level pipelining for computation in one Bundle with four tiles.
Following the top-down approach, parameters of the proposed architecture can be configured to adapt to different FPGA devices and to maximize the performance of FPGA accelerators.

\vspace{-8pt}
\subsection{Bundle and DNN Performance Modeling  \label{sec:modeling}}
\vspace{-2pt}
Based on the proposed \textit{Tile-Arch},
we build analytical models for performance and resource estimation for both Bundles and DNNs used in Bundle evaluation and DNN exploration.
In this work, we take latency as the primary performance measure.

\vspace{-2pt}
\subsubsection{Bundle Performance Modeling}

Denoted a Bundle as $bund_i$, the resource of $bund_i$ is computed as:
\begin{equation}
\label{eq:res-comp}
\vspace{-8pt}
    Res_{bund_i}^r = \sum_{p_j} Res^r_j + \Gamma_i^r
\vspace{-0pt}
\end{equation}
where $Res_j^r$ is the resource usage of instance $p_j$ of resource type $r$ ( including DSP, LUTs, FF and BRAM).
$\Gamma_i^r$ represents other resource overhead such as LUTs consumed by control logic and multiplexers.

The latency of a Bundle is estimated as:
\begin{equation} \label{eq:bund_lat}
\vspace{-8pt}
Lat_{bund_i} = \alpha_i \cdot \sum_{p_j} Comp_j + \frac{\beta_i \cdot \Theta(Data_i)}{bw}
\vspace{-0pt}
\end{equation}
where $Comp_j$ is the computation latency of instance $p_j$, and $\Theta(Data_i)$ is the data amount processed by $bund_i$. 
$bw$ represents the off-chip memory bandwidth.
Denote the latency of one execution of $p_j$ as $lat_j$, and the total number of reuses of $p_j$ as $reuse_j$, the computation latency $Comp_j$ is estimated as:
\begin{equation} \label{eq:bund-comp}
\vspace{-8pt}
    Comp_j = \sum_{1 \leq j \leq n} reuse_{j} \cdot lat_{j}
\vspace{+2pt}    
\end{equation}
$reuse_j$ can be computed by the input/output dimensions of the data processed by the IP and the data dimensions of $p_j$'s interface. The parameter $\alpha_i$ in Eq.~\ref{eq:bund_lat} describes how much computation is overlapped because of IP pipelining,
and $\beta_i$ describes how much data transfer is overlapped during computations.
$\alpha_i$, $\beta_i$ and $\Gamma_i$ will be determined for each $bund_i$ using \textit{Auto-HLS} sampling.

\vspace{-2pt}
\subsubsection{DNN Performance Modeling}
The overall DNN latency based on $Lat_{bund_i}$ in Eq. \ref{eq:bund_lat} is estimated as:
\begin{equation} \label{eq:DNN-overall-latency}
\vspace{-10pt}
    Lat_{DNN} = \sum_{i=1}^{N} Lat_{bund} + \phi \cdot Lat_{DM}
\vspace{+4pt}
\end{equation}
where $N$ is the the number of Bundle repetitions of the DNN, and $\phi \cdot Lat_{DM}$ represents the inter-bundle data movement latency.
For overall DNN resource utilization, we have:
\begin{equation} \label{eq:DNN-overall-resource}
\vspace{-8pt}
    Res_{DNN} = Res_{bund_i} + \gamma \cdot Res_{ctl}
\vspace{+2pt}
\end{equation}
where $Res_{bund_i}$ is the resource of $bund_i$, and $Res_{ctl}$ is additional control logic overhead, e.g., finite state machine and multiplexers.
$\phi$, $\gamma$, $Lat_{DM}$ and $Res_{ctl}$ will be decided through \textit{Auto-HLS} sampling.


\vspace{-4pt}
\section{dnn exploration and update}
\label{sec:auto-dnn}
\vspace{-2pt}

The DNN exploration and update is conducted by \textit{Auto-DNN} cooperated with \textit{Auto-HLS}.
Given a specific machine learning task, coarse- and fine-grained Bundle evaluation is first performed to select
the top-$N$ promising candidates.
After that, a hardware-aware DNN exploration and update is performed,
to search for DNNs within hardware resource and latency constraints.
To better illustrate our approach,
we use an object detection task specified by the 2018 Design Automation Conference System Design Contest (DAC-SDC) \cite{DACSDC} as an example.
This competition targets implementing machine learning applications on an embedded PYNQ-Z1 FPGA (with $4.9$Mbit on-chip BRAM, 220 DSPs, 53,200 LUTs and 106,400 FFs) for board-level designs. 

\vspace{-4pt}
\subsection{Bundle Evaluation and Selection}
\label{sec:bund-eval}
\vspace{-2pt}

\begin{figure}[t]
\centering
\includegraphics[width=0.49\textwidth]{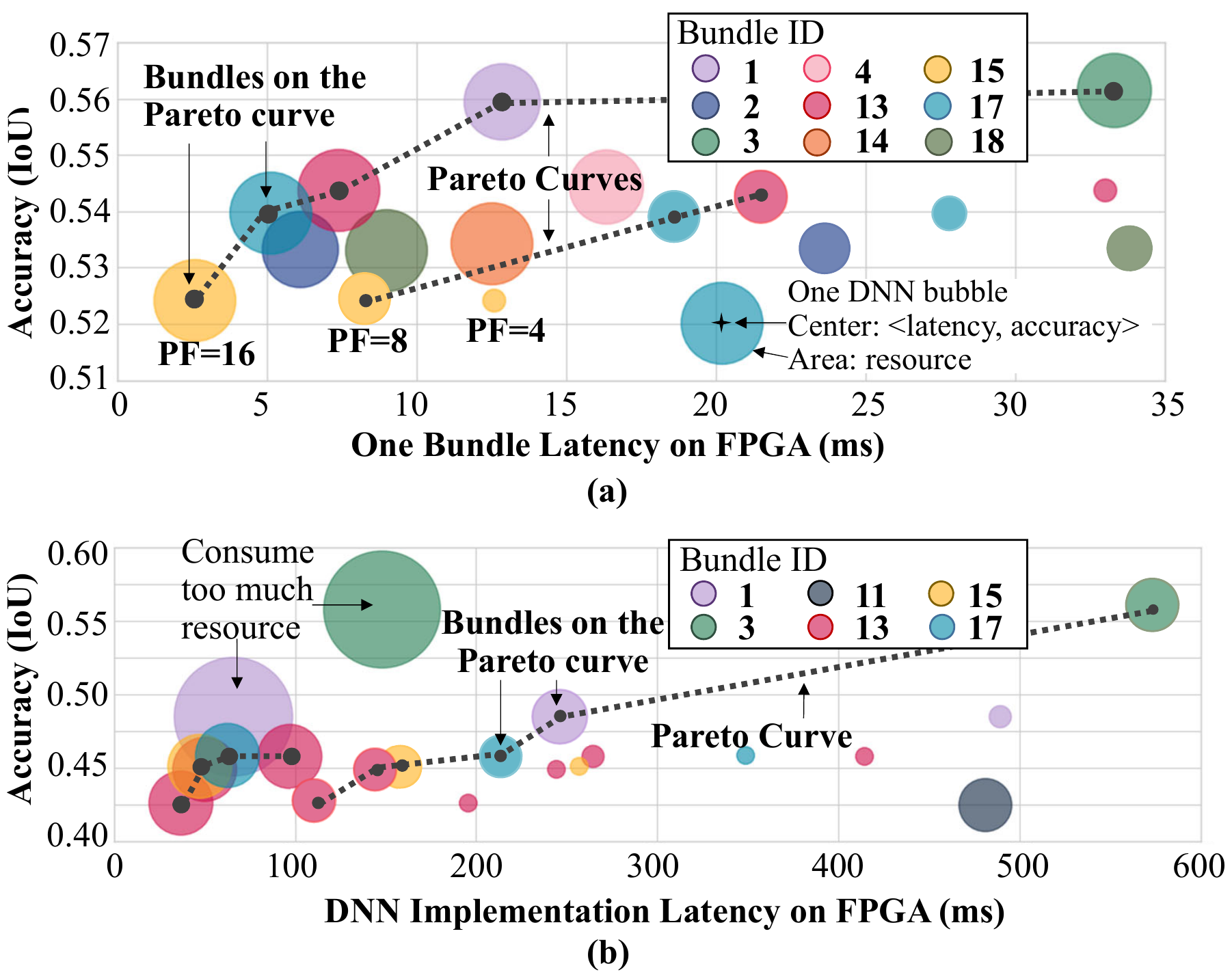}
\vspace{-22pt}
\caption{Coarse-grained bundle evaluation with
(a) DNNs built using \textit{method\#1}; and (b) DNNs built using \textit{method\#2}.
}
\label{fig:coarse_bund_eval}
\end{figure}

\begin{figure}[t]
\centering
\includegraphics[width=0.45\textwidth]{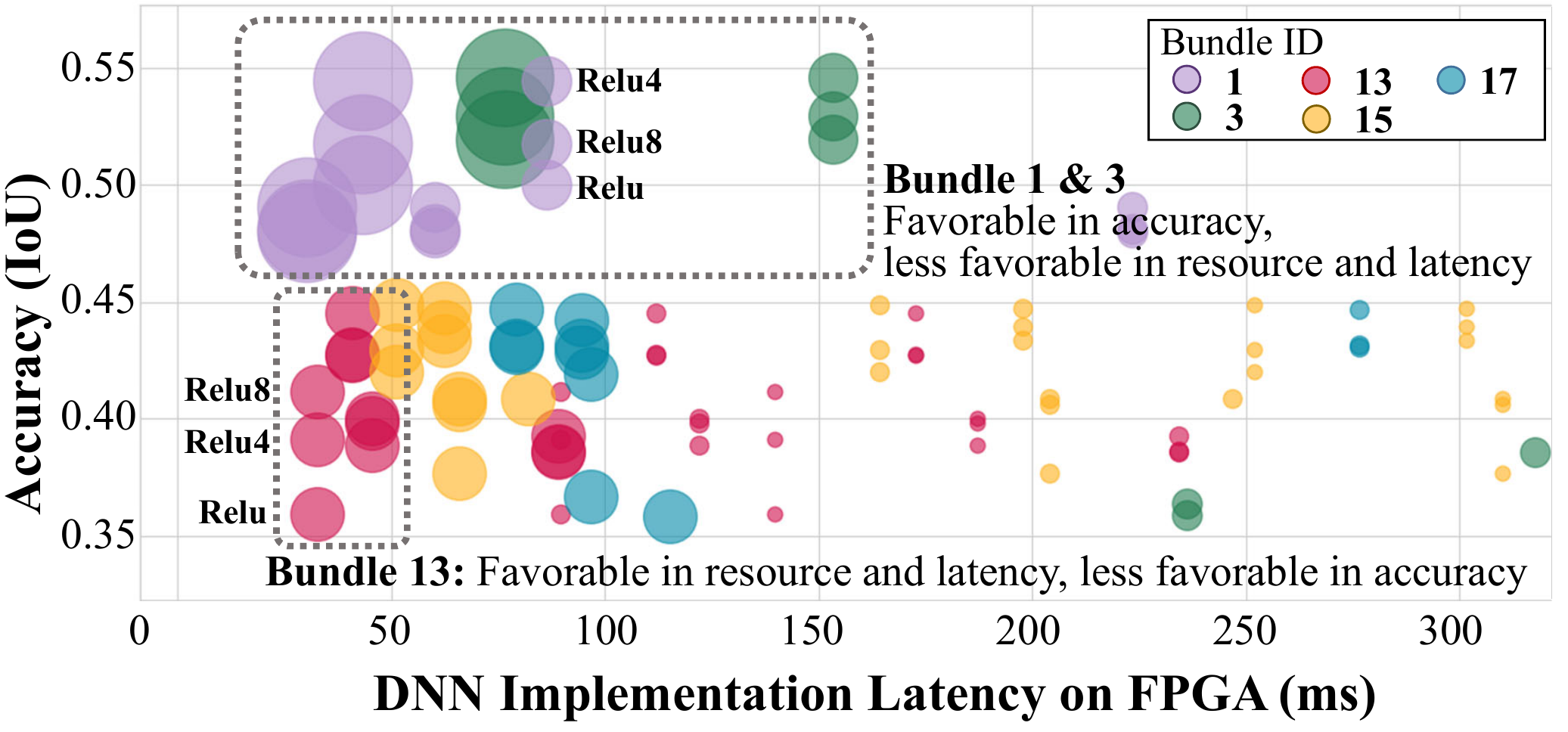}
\vspace{-12pt}
\caption{Fine-grained evaluation of the selected Bundles.}
\vspace{-6pt}
\label{fig:bundle_fine_eval}
\end{figure}

\subsubsection{Coarse-Grained Evaluation}
\label{sec:coarse-bund-eval}
In this step, a three-dimensional feature including latency, resource and accuracy is captured for each Bundle.
For latency and resource, we use Bundle and DNN modeling in Sec.~\ref{sec:modeling}; for accuracy, we train the DNNs built by Bundles on the target dataset.
This evaluation is critical for co-design scalability,
especially when a large number of Bundle candidates are provided for complex machine learning tasks.

We propose two methods to construct DNNs to evaluate Bundle accuracy.
\textit{method\#1}: we use a DNN template with a fixed head and tail, and insert one Bundle replication in the middle;
\textit{method\#2}: we replicate a Bundle for $n$ times to build a DNN.
Since Bundles may perform differently on various machine learning tasks, the constructed DNNs are directly trained on the target task in a \textit{proxyless} manner \cite{cai2018proxylessnas}.
For fast evaluation, each DNN is trained for a small number of epochs (20 in the experiment).
After evaluation, Bundles with similar resource usage (e.g. DSPs) are grouped,
and a Pareto curve is generated for each group. The Bundles on the Pareto curve will be selected.

Fig.~\ref{fig:coarse_bund_eval} illustrates the coarse bundle evaluation on the example object detection task.
Each bubble represents a DNN built from a Bundle.
The coordinates of the bubble center represent latency and accuracy of the DNN, while the area of the bubble represents the resource usage.
Under different parallel factors (PF), the implementation of a DNN differs in latency and resource but has the same accuracy.
From Fig.~\ref{fig:coarse_bund_eval} (a) and (b), we notice that both methods of constructing DNNs can deliver similar results,
where the Bundles on the Pareto curve are the same from both curves (Bundle 1, 3, 13, 15 and 17).
It implies that our proposed Bundle evaluation is reliable for Bundle selection.

\vspace{-2pt}
\subsubsection{Fine-Grained Evaluation}
\label{sec:find-bund-eval}

After coarse-grained evaluation, a fine-grained evaluation on the selected Bundles is performed
to better understand their characteristics. 
We construct DNNs by replicating certain Bundles for $n$ times,
and also try different activation functions such as $Relu4$ and $Relu8$, which relate to data quantization.
Fig.~\ref{fig:bundle_fine_eval} shows the fine-grained evaluation results for selected Bundles.
It reveals that each Bundle has its own characteristics regarding latency, accuracy and resource overhead.
For example, Bundle 1 and 3 are more promising in high accuracy DNNs with more resource and longer latency, while Bundle 13 is more favorable in DNNs targeting real-time responses with less resource.

\vspace{-8pt}
\subsection{Hardware-Aware DNN Search and Update}
\vspace{-2pt}

After selecting top-$N$ promising Bundle candidates,
\textit{Auto-DNN} searches DNN models under resource and latency constraints.
For each Bundle, $K$ initial DNNs are generated and are incrementally updated until the latency target is met.
Inside \textit{Auto-DNN}, a Stochastic Coordinate Decent (SCD) unit is used for DNN update.

\vspace{-2pt}
\subsubsection{DNN Initialization}
For each $bund_i$, total $K$ DNNs will be generated, trained and fine-tuned as outputs.
Output DNNs are denoted as $DNN_i^{k}$ ($1 \leq k \leq K$),
and each starts from an initial one denoted as $DNN_i^{k_0}$.
First, we initialize software related variables.
The $bund_i$ is replicated with $N_i$ times;
initial down sampling layers are inserted between replications;
initial channel expansion factors are set to be 1 (do not expand) or 2 (double the number of channels), depending on the layer type.
Next, hardware related variables will be traversed.
Given $bund_i$, the IP templates, i.e., 
the $IP_1$ to $IP_m$ in Table \ref{tab:co-design-var}, are determined, and $p_1$ to $p_m$ are instantiated.
For simplicity, each IP template is instantiated into one $p_j$,
configured with parallel factor $PF_j$ and quantization scheme $Q_j$.
We let $Q_j$ and $PF_j$ to be consistent among all IP instances to allow IP reuse across layers and BRAM buffer reuse across IPs.
Under a certain $Q_j$, $PF_j$ is set as the maximum value that can fully utilize available resources.

\vspace{-4pt}
\subsubsection {Stochastic Coordinate Descent (SCD) Unit} \label{sec:dnn-update}

The SCD unit takes an initial $DNN_i^{k_0}$ as its input,
together with a latency target $L_{targ}$, latency tolerance $\epsilon$,
and resource constraint $Res_{max}$.
Denote the achieved latency of $DNN_i^{k}$ as $Lat$ and achieved resource as $Res$,
the objective of SCD unit is $|Lat_{targ} - Lat| < \epsilon$ and $Res < Res_{max}$.

The SCD procedure is shown in Algorithm \ref{alg:DNN-search}.
Given an initial $DNN_i^{k_0}$, 
the SCD algorithm updates three variables:
the number of Bundle replications, denoted as $N_i$;
down-sampling configurations between bundles, denoted as $X$, which is a vector with zero-one entries indicating without/with down-samplings between Bundles;
channel expansion configuration, denoted as $\Pi$, representing the vector $<f_{ch_1}, \cdots>$ in Table~\ref{tab:co-design-var}.
The available channel expansion factors include $\{1.2, 1.3, 1.5, 1.75, 2\}$.
Denote a unit \textit{move} as $\Delta$, the moves along three coordinates as $\Delta_{N}$, $\Delta_{\Pi}$ and $\Delta_{X}$,
and the latency changes because of the moves as $\Delta Lat_{N}$, $\Delta Lat_{\Pi}$ and $\Delta Lat_{X}$, respectively.
Given the difference between $Lat_{targ}$ and $Lat$ 
as $\Delta L=|Lat_{targ}-Lat|$,
the number of unit moves along $N$, $\Pi$ and $X$ directions are computed as $\Delta L/\Delta Lat_{N}$, $\Delta L/\Delta Lat_{\Pi}$ and $\Delta L/\Delta Lat_{X}$.
Then, the SCD algorithm picks one coordinate in random, and updates $DNN_i^k$ along that direction within resource constraints.

When the objective of SCD is met, $DNN_i^k$ is saved into set $DNN_s$ as a candidate DNN.
The $K$ candidates are passed to DNN training framework to get their accuracy.
Meanwhile, the DNNs are also passed to \textit{Auto-HLS} to generate their FPGA implementations
and get synthesized resource usage and latency.

\subsubsection{Auto-HLS}
To automatically generate FPGA accelerators for DNNs helps reduce the FPGA development cycle and engineering hours.
Following the \textit{Tile-Arch} template,
\textit{Auto-HLS} generates C code for FPGA accelerators,
which can be directly synthesized by HLS tools.
Since our IPs are written in C, knowing the input/output data dimensions of each IP and feature maps, the \textit{Auto-HLS} generates function calls for the IPs with corresponding weight loading and data buffering functions.
After C code generation, manual optimizations may be applied such as buffer re-allocation and loop fusion, which will be automated in the near future.



\begin{algorithm}[t]
\footnotesize

\caption{DNN Exploration with Stochastic Coordinate Decent\vspace{-2pt}}\label{alg:DNN-search}
\begin{algorithmic}[1]
\Require $Lat_{targ}$, $Lat$ tolerance $\epsilon$, $Res_{targ}$, initial $DNN_i^{k_0}$
\Ensure $K$ DNNs s.t. $|Lat_{targ} - Lat| < \epsilon$, $|Res < Res_{max}|$

\State Selected DNNs: $DNNs \leftarrow \emptyset$, initialize $N$, $\Pi, X\leftarrow DNN_i^{k_0}$

\While{$k < K$}
    \State $Lat \leftarrow$\textsc{Est\_Lat}($DNN_i^k$)
    \If{$|Lat_{targ} - Lat| < \epsilon$}
        \State $k\leftarrow k+1$, $DNNs \leftarrow DNNs \cup DNN_i^k$
    \EndIf

        \State $\Delta Lat_{N} \leftarrow $\textsc{Est\_Lat}($(DNN[i_{N} + \Delta N ])$)$-Lat$
        \State $\Delta Lat_{\Pi} \leftarrow $\textsc{Est\_Lat}($(DNN[i_\Pi + \Delta \Pi ])$)$-Lat$
        \State $\Delta Lat_{X} \leftarrow $\textsc{Est\_Lat}($(DNN[i_{X} + \Delta X ])$)$-Lat$
    
        \State Pick $\Delta \leftarrow \{ \Delta N, \Delta \Pi, \Delta X\}$ uniformly at random

    \If { $\textsc{Est\_Res}((DNN[i + \Delta ])) < Res_{max} $ }
    
        \State \textbf{if} $\Delta = \Delta N $ \textbf{then} $\Delta_{N} \leftarrow \lfloor|Lat_{targ} - Lat| / \Delta Lat_{N} \rfloor$, $i_{N} \leftarrow i_{N}+\Delta_{N}$
        \State \textbf{if} $\Delta = \Delta \Pi $ \textbf{then}  $\Delta_{\Pi} \leftarrow \lfloor|Lat_{targ} - Lat| / \Delta Lat_{\Pi} \rfloor$, $i_\Pi \leftarrow i_\Pi+\Delta_{\Pi}$
        \State \textbf{if} $\Delta = \Delta X $ \textbf{then}  $\Delta_{X} \leftarrow \lfloor|Lat_{targ} - Lat| / \Delta Lat_{X} \rfloor$, $i_X \leftarrow i_X+\Delta_{X}$
    
    \EndIf
    
    \State $DNN_i^k \leftarrow (DNN[i_{N}, i_\Pi, i_X])$
\EndWhile
\\
\Return $DNNs$
\end{algorithmic}
\end{algorithm}

\begin{table*}[t]
\footnotesize
\centering
\caption{Performance Comparisons (FPGA and GPU competition data are obtained from \cite{xu2018dacsdc}) \vspace{-12pt}} \label{tab:res}
\renewcommand{\arraystretch}{0.95}
\setlength{\tabcolsep}{6pt}
\begin{tabular}{|c | c | c c c c c c | c c c c|}
\hline 
                         & \multirow{2}{*}{Model}  & \multirow{2}{*}{IoU} & \multirow{2}{*}{Latency}
                         & \multirow{2}{*}{FPS} & \multirow{2}{*}{Power} & \multirow{2}{*}{Energy}
                         & \multirow{2}{*}{Efficiency} & \multicolumn{4}{|c|}{Resource Utilization} \\ \cline{9-12}

                         &				 &     &       &  &     &       &	& LUTs & DSP & BRAM & FF \\
\hline \hline
\multirow{6}{*}{Ours} 	 & DNN1 & 68.6\% & 80.0 ms (100 MHz) & 12.5 & 2.2W & 8.80 KJ & 0.18 J/pic & 82.5\% & 91.8\% & 96.1\% & 37.6\%\\
 						 & 					     &        & \textbf{57.4 ms (150 MHz)} & \textbf{17.4} & \textbf{2.5W} & \textbf{7.18 KJ} & \textbf{0.14 J/pic}  & 82.5\% & 91.8\% & 96.1\% & 37.6\%\\
  						 & DNN2 & 61.2\% & 62.6 ms (100 MHz) & 16.0 & 2.2W & 7.50 KJ & 0.15 J/pic  & 76.4\% & 84.6\% & 77.9\% & 27.4\%\\
                         &                       &        & 44.1 ms (150 MHz) & 22.7 & 2.4W & 5.51 KJ & 0.11 J/pic  & 76.4\% & 84.6\% & 77.9\% & 27.4\%\\
  						 & DNN3 & 59.3\% & 47.8 ms (100 MHz) & 20.9 & 2.2W & 5.74 KJ & 0.11 J/pic  & 70.4\% & 85.2\% & 95.4\% & 32.2\%\\
                         &                       &        & \textbf{33.7 ms (150 MHz)} & \textbf{29.7} & \textbf{2.4W} & \textbf{4.04 KJ} & \textbf{0.08 J/pic}  & 70.4\% & 85.2\% & 95.4\% & 32.2\%\\

\hline
1st in FPGA 	 &  SSD & 62.4\% & 84.6 ms (150 MHz) & 11.96 & 4.2W & 17.56 KJ & 0.35 J/pic  & 83.9\% & 100\% & 78.9\% & 54.2\%\\
2nd in FPGA 	 & -- & 49.2\% & 38.5 ms (150 MHz) & 25.97 & 2.5W & 4.81 KJ & 0.10 J/pic  & 88\% & 78\% & 77\% & 62\%\\
3rd in FPGA 	 & -- & 57.3\% & 136.1 ms (150 MHz) & 7.35 & 2.6W & 17.69 KJ & 0.35 J/pic  & 63\% & 86\% & 95\% & 22\%\\

\hline
1st in GPU 	 & Yolo & 69.8\% &  40.7 ms (854 MHz) & 24.55 & 12.6W & 25.66 KJ & 0.51 J/pic 
& - & - & - & -\\
2nd in GPU 	 & Tiny-Yolo & 69.1\% & 39.5 ms (854 MHz) & 25.3 & 13.3W  & 26.28 KJ & 0.53 J/pic 
& - & - & - & -\\
3rd in GPU 	 & Tiny-Yolo & 68.5\% & 42.3 ms (854 MHz) & 23.64 & 10.3W & 21.79 KJ & 0.44 J/pic 
& - & - & - & - \\
\hline 
\end{tabular}
\end{table*}

\vspace{-4pt}
\section{experimental results \label{sec:result}}
\vspace{-2pt}

\begin{figure}[t]
\centering
\includegraphics[width=0.45\textwidth]{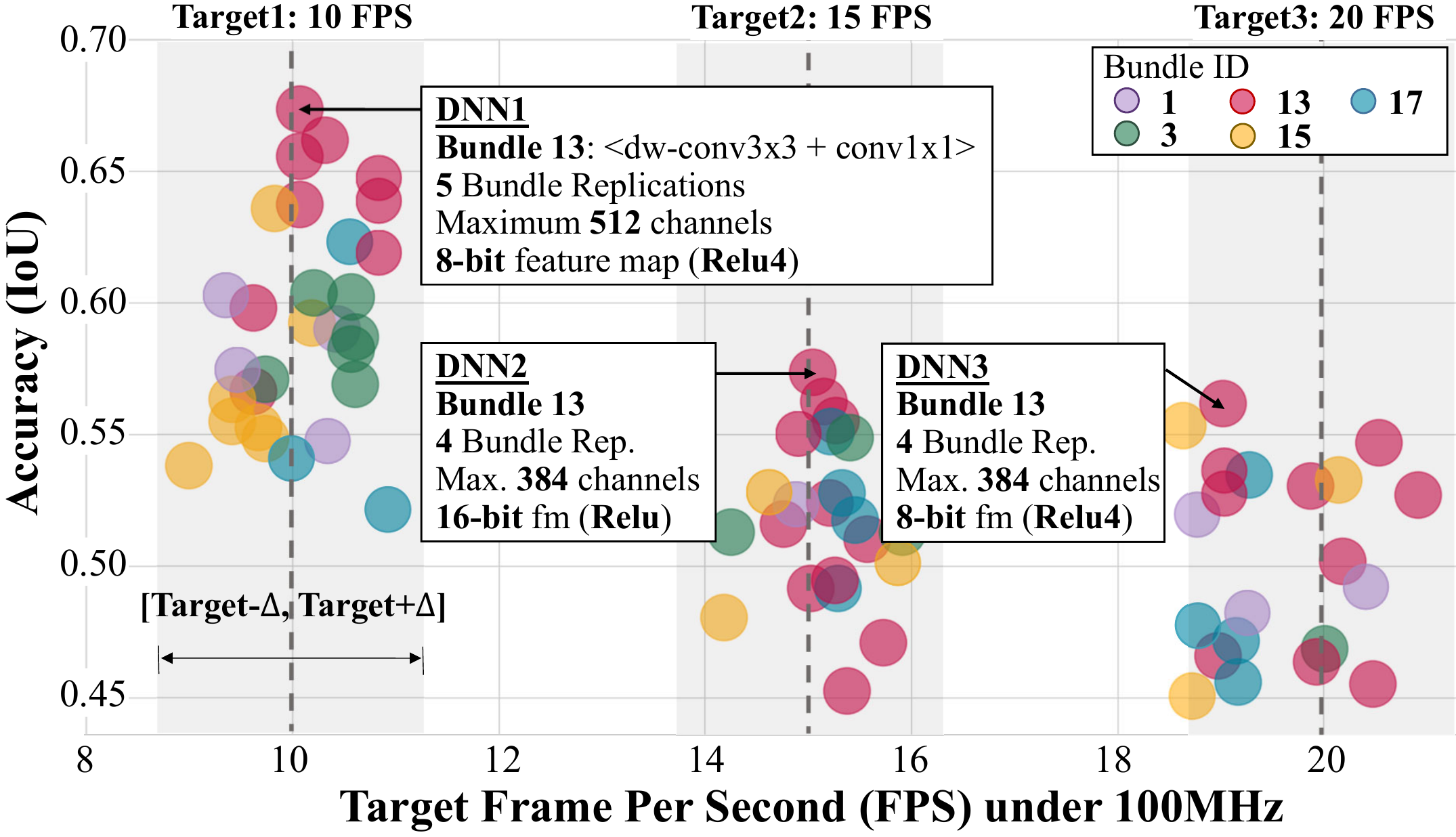}
\vspace{-8pt}
\caption{DNN models explored targeting 10/15/20 FPS @ 100MHz.}
\vspace{-12pt}
\label{fig:finalrst}
\end{figure}

For demonstration, we use the same object detection task as in Sec.~\ref{sec:auto-dnn}.
To provide trade-off options between DNN latency and accuracy, we set three latency targets: 10, 15 and 20 FPS at 100MHz.

By specifying the resource constraints and latency targets,
our proposed co-design methodology 
conducts DNN model exploration using selected Bundles,
and outputs DNNs with their corresponding accelerators.
Fig. \ref{fig:finalrst} shows all the explored DNNs that meet target latency within resource constraints.
The DNNs which fall into the range $[target-\Delta,target+\Delta]$ are considered as candidates output for training.
In total, 68 DNN models are built from 5 different Bundles with training and fine-tuning. Among them, we pick those with the best accuracy for each FPS target and get DNN1$\sim$3.
The detailed structures of the final DNNs are shown in
Fig.~\ref{fig:finalrst}.
DNN1 achieves the highest IoU, reaching 68.6\% with 12.5 FPS@100MHz and 17.4 FPS@150MHz. 
DNN2 achieves 61.2\% IoU with 16.0 FPS@100MHz and 22.7 FPS@150MHz, while
DNN3 achieves the highest FPS at 29.7 FPS@150MHz with 59.3\% IoU. 
Some additional modifications are applied on the \textit{Auto-HLS} generated C code, such as on-chip buffer allocation and loop fusion, to reach higher FPS. 

\begin{figure}
  \centering
  \vspace{-0pt}
  \includegraphics[width=0.46\textwidth]{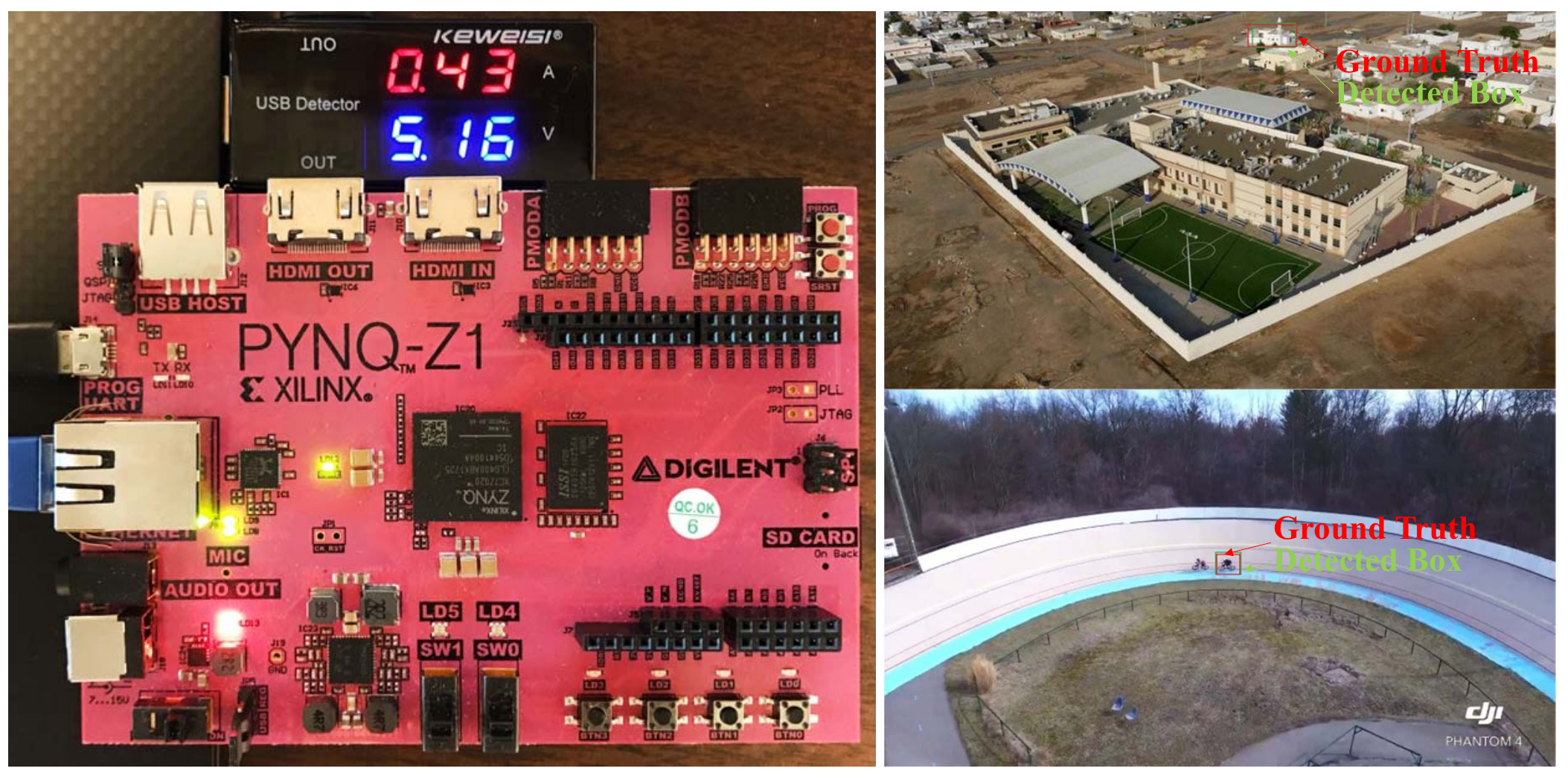}
  \vspace{-12pt}
  \caption{Pynq-Z1 board with power meter measured while running object detection.}\label{fig:board}
  \vspace{-20pt}
\end{figure}

We also compare to the state-of-the-art works for this object detection task on PYNQ-Z1 published in~\cite{xu2018dacsdc}.
The comparisons to FPGA and GPU categories are shown in Table \ref{tab:res}.
The results are collected from the board-level implementations. The IoU is measured on 50K images from the official dataset following the same criteria in DAC-SDC. Latency refers
to a single frame latency, while FPS is measured using total run-time for the 50K images including image loading, preprocessing, and DNN inference. The power and energy are measured using the POWER-Z KT001 USB Power Monitor as shown in Fig. \ref{fig:board}. We also show two example images with the ground truth bounding boxes (red) and our generated boxes (green).

Compared to the 1st-place winner of the FPGA category, we achieve 6.2\% higher IoU, 40\% lower power, and 2.5$\times$ better energy efficiency. 
The 1st-place FPGA team follows the top-down design flow by starting from a standard DNN-based detector (SSD). After network compression, the DNN is small enough that satisfies both hardware constraints and performance demands \cite{1stFPGA}. Compared to this top-down approach, our co-design method is able to deliver better DNN models and more efficient hardware accelerators.
Compared to GPU-based designs, our DNN1 model is more accurate than the 3rd-place design and only 1.2\% lower IoU than the 1st-place GPU design. 
Regarding the energy efficiency, ours is 3.6$\times$ better than the 1st-place GPU design with 40\% longer latency despite a nearly 6$\times$ slower clock frequency.

\vspace{-8pt}
\section{Conclusion}
\vspace{-4pt}
We presented an FPGA/DNN co-design methodology with both bottom-up DNN model exploration and top-down accelerator design approaches to enhance the IoT intelligence on embedded FPGAs.
On the defined co-design space, we proposed \textit{Auto-DNN}, an automatic DNN model search engine to explore hardware-friendly DNNs, and an automatic HLS generator, \textit{Auto-HLS}, to generate FPGA-based DNN accelerators.
We applied our proposed methodology to an object detection task from DAC-SDC competition. Results showed that our implementation outperformed the 1st place winner in all factors with 6.2\% higher IoU, 40\% lower power, and 2.5$\times$ better energy efficiency. Comparing to GPU designs, our results achieved similar accuracy (0.1\% better than 3rd place and 1.2\% worse than 1st place) but with 3.1$\times$ to 3.8$\times$ better energy efficiency.

\vspace{-6pt}
\begin{acks}
\vspace{-4pt}
  This work was partly supported by the IBM-Illinois Center for Cognitive Computing System Research (C$^3$SR) -- a research collaboration as part of IBM AI Horizons Network.
\vspace{-6pt}
\end{acks}

%

%
\bibliographystyle{unsrt}
\bibliographystyle{ACM-Reference-Format}
\bibliography{sample-sigconf}

\end{document}